\def\year{2021}\relax
\documentclass[letterpaper]{article} % DO NOT CHANGE THIS
\usepackage{aaai21}  % DO NOT CHANGE THIS
\usepackage{times}  % DO NOT CHANGE THIS
\usepackage{helvet} % DO NOT CHANGE THIS
\usepackage{courier}  % DO NOT CHANGE THIS
\usepackage[hyphens]{url}  % DO NOT CHANGE THIS
\usepackage{graphicx} % DO NOT CHANGE THIS
\usepackage{multirow}
\usepackage{colortbl}
\usepackage{hhline}
\usepackage{amsfonts}
\usepackage{amssymb,amsmath}
\usepackage{natbib}  % DO NOT CHANGE THIS AND DO NOT ADD ANY OPTIONS TO IT
\usepackage{caption} % DO NOT CHANGE THIS AND DO NOT ADD ANY OPTIONS TO IT
\usepackage[inline]{trackchanges} % change inline to finalnew to accept all changes
\addeditor{DW}
\addeditor{WQ}
\addeditor{TM}

\title{Explore the Potential Performance of Vision-and-Language Navigation Model: a Snapshot Ensemble Method}
\author{
    Wenda Qin\textsuperscript{\rm 1}, Teruhisa Misu\textsuperscript{\rm 2}, Derry Wijaya\textsuperscript{\rm 1} \\
}
\affiliations{
    \textsuperscript{\rm 1}Boston University\\
    111 Cummington Mall, Boston MA 02215 \\
    wdqin, wijaya@bu.edu \\
    
    \textsuperscript{\rm 2}Honda Research Institute USA, Inc. \\
    70 Rio Robles, San Jose, CA 95134 \\
    tmisu@honda-ri.com \\

}
\begin{document}

\maketitle

\begin{abstract}
Vision-and-Language Navigation (VLN) is a challenging task in the field of artificial intelligence. Although massive progress has been made in this task over the past few years attributed to breakthroughs in deep vision and language models, it remains tough to build VLN models that can generalize as well as humans. In this paper, we provide a new perspective to improve VLN models. Based on our discovery that snapshots of the same VLN model behave significantly differently even when their success rates are relatively the same, we propose a snapshot-based ensemble solution that leverages predictions among multiple snapshots. Constructed on the snapshots of the existing state-of-the-art (SOTA) model $\circlearrowright$BERT \textit{and} our past-action-aware modification, our proposed ensemble achieves the new SOTA performance in the R2R dataset challenge in Navigation Error (NE) and Success weighted by Path Length (SPL).
%Vision-and-Language Navigation (VLN) has been one of the most challenging tasks for the artificial intelligence community since its birth. Massive progress has been made in the VLN task over the past few years, attributed to breakthroughs in deep vision and language models. However, building VLN models that generalize as well as human still has a long way to go.

%In this paper, we provide a new perspective to improve the VLN models. We discovered that snapshots of the same VLN model behave significantly differently while their success rates are relatively the same. In order to take advantage of snapshots with different behaviors, we propose a snapshot-based ensemble solution to leverage the predictions among multiple snapshots. Based on a Recurrent PREVALENT model and a past-action-aware modification, our snapshot ensemble achieved the new SOTA performance in the R2R dataset challenge in Navigation Error (NE) and Success weighted by Path Length (SPL).
\end{abstract}

\section{Introduction}
\noindent Given instruction in texts and visual inputs describing the surrounding environment, a Vision-and-Language Navigation (VLN) agent controls an agent to complete a set of goals listed in the instruction. Building a good VLN system is difficult due to the fact that it needs to understand vision and language information and coordinate them well. %like humans, which is impossible for systems in the past.

Recent advancements in computer vision and natural language processing and the advent of better vision-language models \cite{sundermeyer2012lstm,vaswani2017attention,lu2019vilbert,tan2019lxmert} along with the effort to prepare large scale realistic datasets \cite{Matterport3D} has enabled rapid development of VLN systems. 
%There's rapid development related to VLN systems recently. On the one hand, people are now able to build much better language and vision models \cite{sundermeyer2012lstm,vaswani2017attention,lu2019vilbert,tan2019lxmert}. On the other hand, more and more datasets based on realistic scenes have become available \cite{Matterport3D}, providing more potentials to the VLN system.
Among all current VLN datasets, the R2R dataset (Anderson et al. 2018) is a dataset based on real photos taken in indoor environments. It attracts massive attention for its simple-form task, which at the same time requires complex understanding in both images and texts.

To obtain better performance in R2R, various studies in the past have discussed how to adjust the best vision/language models at the time for the R2R VLN task \cite{anderson2018vision,majumdar2020improving,hong2021vln}. Previous studies have also made efforts to prevent overfitting due to the limited size of the R2R dataset \cite{fried2018speaker,liu2021vision,li2019robust,hao2020towards}. 

In this paper, we offer a new perspective for analyzing the R2R VLN model that focuses on the by-products of the model training process: snapshots. Snapshots are the saved parameters of a model at various intervals during training. Although all snapshots have the same goal as the model, their parameters are different due to the ongoing optimization. We discover that some of the best snapshots at various intervals saved during training shared similar navigation success rates while making significantly diverse errors. Based on such observation, we construct our VLN system with an ensemble of snapshots instead of just one. Through experiments, we found out that such an ensemble can take advantage of its members and thus significantly improve the navigation performance. 

In addition, to allow more model variants in the ensemble, we also propose a novel modification of an existing state-of-the-art (SOTA) model for VLN i.e., the VLN$\circlearrowright$BERT \cite{hong2021vln}. Our ensemble, which consists of snapshots of both models: the VLN$\circlearrowright$BERT model \textit{and} our proposed modification: the past-action-aware VLN$\circlearrowright$BERT model--achieves a new SOTA performance in the single-run setting of the R2R dataset. 
%modified version of the State-Of-The-Art (SOTA) model. Based on the two versions of the model we have, our ensemble achieved a new SOTA performance in the single-run setting of the R2R dataset. 

To conclude, our contributions are as follows:

\begin{itemize}
\item We discover that the best snapshots of the same model behave differently while having similar navigation success rates. Based on this observation, we propose a snapshot ensemble method to take advantage of the different snapshots. %act differently while their success rates are relatively close. According to the above observation, We propose a snapshot ensemble method to take advantage of the difference among snapshots.
\item We also propose a past-action-aware modification on the current best VLN model: the VLN$\circlearrowright$BERT. It creates additional variant snapshots to the original model with equivalent navigation performance.
\item By combining the snapshots from both the original and the modified model, our ensemble achieves a new SOTA performance on the R2R challenge leaderboard in the single-run setting.\footnote{Our method is noted as ``SE-Mixed (Single-Run)" in the leaderboard webpage: https://eval.ai/web/challenges/challenge-page/97/leaderboard}
\item We evaluate the snapshot ensemble method on two different datasets and apply it with two other VLN models. The evaluation results show that the snapshot ensemble also improves performance on more complicated VLN tasks and with different model architectures.
\end{itemize}

\section{Related Works}
\subsection{Vision-and-language Navigation datasets}
Teaching a robot to complete instructions is a long-existing goal in the AI community \cite{winograd1971procedures}. Compared to GPS-based navigation, VLN accepts surrounding environments as visual inputs and correlates them with instruction in human language. Most VLN datasets in the past are based on synthesized 3-D scenes \cite{kolve2017ai2,brodeur2017home,wu2018building,yan2018chalet,Song_2017_CVPR}. Recently, the emergence of data based on real-life scenarios allows VLN systems to be developed and tested in realistic environments. Specifically, 3-D views from Google Street View\footnote{\url{https://www.google.com/streetview/}} and Matterport3D datasets (Chang et al. 2017) allow people to build simulators that generate navigation data from photos taken in real life. Different from the previous datasets, the R2R dataset that we use consists of navigations in real indoor environments. Concretely, the R2R dataset provides $\sim$ 15,000 instructions and $\sim$ 5,000 navigation paths in 90 indoor scenes. Since its construction, people have proposed variants of the R2R dataset to address certain shortcomings of the original one \cite{ku2020room,jain2019stay,hong2020sub,krantz2020beyond}. However, the community still considers the R2R dataset a necessary test for evaluating all kinds of VLN systems for indoor navigation. 

\subsection{VLN systems for navigation in R2R dataset}
To improve navigation performance in the R2R dataset, various models and techniques have been proposed. \citet{fried2018speaker} and \citet{tan2019learning} further developed the LSTM \cite{sundermeyer2012lstm} + soft-attention \cite{luong2015effective} baseline system \cite{anderson2018vision}. \citet{majumdar2020improving} proposed a VLN system based on VilBERT \cite{lu2019vilbert} to replace the LSTM + soft-attention architecture for better image and text understanding. Recently, \citet{chen2021topological,Wang_2021_CVPR,hong2020language} proposed VLN systems based on graph models. In terms of techniques, \citet{fried2018speaker} built a speaker model for data augmentation. \citet{ma2019regretful,ma2019self} introduced regularization loss and back-tracking; \citet{tan2019learning} improved the dropout mechanic in its VLN model; \citet{li2019robust,hao2020towards} improved the models' initial states by pre-training it on large-scale datasets; and \citet{hong2021vln} developed a recurrent VLN model (VLN$\circlearrowright$BERT) based on BERT structure for single-run setting in VLN. \citet{liu2021vision} provides further data augmentation by splitting and mixing scenes.

Previous work that shared the closest idea to us is \citet{hu2019you}, which proposed a mixture of VLN models. However, each of their models is trained with different inputs. In this paper, we build an ensemble based on snapshots of the \textit{same} model.

\subsection{Ensemble}
The concept of applying ensemble in neural network models appeared very early in the machine learning community \cite{hansen1990neural}. There are well-known ensemble techniques such as Bagging \cite{breiman1996bagging}, Random forests \cite{ho1995random}, and boosting (AdaBoost) \cite{freund1997decision}. However, applying such ensembles to deep learning models directly is very time-consuming. There are previous works that provide ensemble-like solutions for deep learning models \citet{xie2013horizontal,moghimi2016boosted,laine2016temporal,french2017self}. Our work is inspired by the idea of ``snapshot ensemble'' from \citet{huang2017snapshot}, which constructs the ensemble from a set of snapshots collected in local minima. Different from the previous work, we collect snapshots based on training intervals and success rates. Also, we apply beam search to optimize the combination of snapshots to be in the ensemble.

\section{Preliminaries}
% In this section, we provide the definition of the VLN task in R2R, the VLN model we apply snapshot ensemble to, and the observation we had on the model that led us to the snapshot ensemble solution.
\subsection{Vision-and-language Navigation in R2R dataset}
Navigation in R2R consists of three parts: instruction $I$, scene $S$, and path $P$.
The instruction $I$ is a sequence of $L$ words in the vocabulary $W$: $I = \{w_1, w_2, ..., w_L\ | w_i \subset W, 1\leq i \leq L \}$. The scene $S$ is a connected graph that contains viewpoints $V$ and the edges $E$ that connect viewpoints: $S=\{V, E\}$. For viewpoint $v_i \subset V$ where the agent stands, there's is a panoramic view $O_i$ to describe the visual surroundings of the agent. To be more precise, $O_i$ is a set of 36 views $O_i = \{o_{i,j}\}^{36}_{j=1}$ that a camera captured in the viewpoint from different horizontal and vertical directions.  A viewpoint $v_a$ is connected (``navigable") to another viewpoint $v_b$ when you could directly walk from $v_a$ to $v_b$ in the real environment that $S$ represents. The path $P$ is a sequence of viewpoints, which starts from the starting viewpoint, and ends in the destination viewpoint: $P = \{v_1,v_2,...,v_n | v \subset V \}$. $v_1$ is the initial position of the agent.

A VLN model for the R2R navigation task works as a policy function with the instruction $I$ and the panoramic view $O_i$ of a certain viewpoint as inputs: $\pi_{\theta}(I, O_i)$. At each time step $t$, the policy function predicts an action $a_t \sim \pi_{\theta}(I,O_i)$, and tries to get as close as possible to the ground truth destination $v_n$ in $P$ at the end.

After the agent chooses to stop or the number of its actions exceeds a limit, its last viewpoint $v_{end}$ will be evaluated. If $v_{end}$ is within 3 meters of $v_n$ from ground-truth path $P$, the navigation is considered to be successful or failed otherwise.

There are three different settings for the VLN task in R2R: single-run, pre-explore, and beam search. In this paper, we focus on the single-run setting. The ``single-run" setting requires the agent to finish the navigation with minimum actions taken and without prior knowledge of the environment.

\subsection{VLN$\circlearrowright$BERT model}
\begin{figure}[ht]
\centering
\includegraphics[width=0.9\columnwidth]{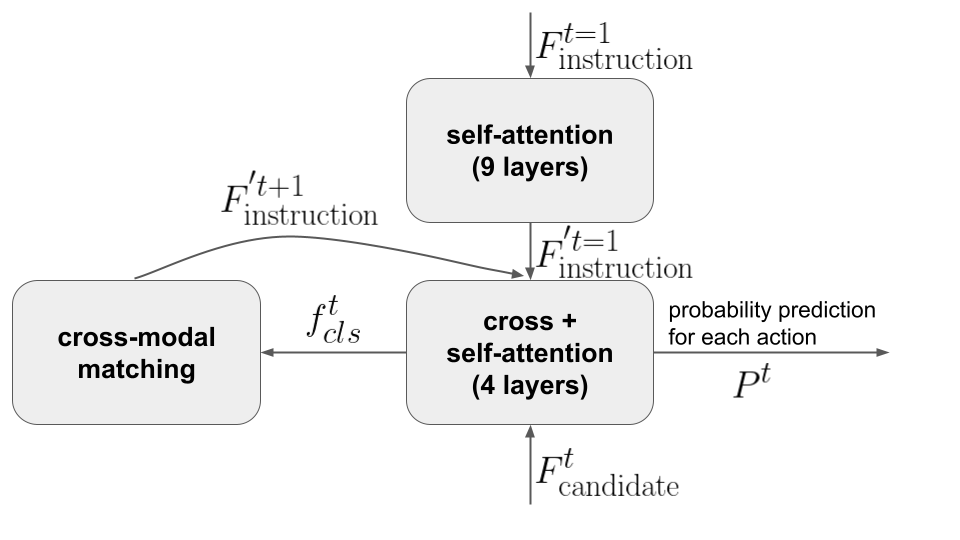}
\caption{A visualization of the VLN$\circlearrowright$BERT %Recurrent-PREVALENT 
model. The instruction feature first passes through a self-attention module and then attends to a candidate feature vector through a cross-self-attention module. The candidate feature then self-attended itself in the same module. After four layers of computation, the last layer outputs the probabilities of each action and sends the $cls$ feature to a cross-modal matching module. The output will replace the $cls$ features in the instruction vector for the next time step.} 
\label{fig_recurrentprevalent}
\end{figure}

We apply our modification and snapshot ensemble on the VLN$\circlearrowright$BERT model proposed by \citet{hong2021vln}. \footnote{\cite{hong2021vln} proposed two VLN$\circlearrowright$BERT models in their work. The VLN$\circlearrowright$BERT model we used here is the LXMERT-based VLN$\circlearrowright$BERT model pre-trained by PREVALENT \cite{hao2020towards}. For the other one, which is BERT-based and pre-trained by OSCAR \cite{li2020oscar}, we call it OSCAR-init VLN$\circlearrowright$BERT to distinguish it from the PREVALENT-initialized one.} 
The model currently holds the best performance for the single-run setting in the R2R dataset \cite{liu2021vision}. In this section, we will have a brief recap of this model. %the Recurrent-PREVALENT model. 
A simplified visualization of the model %Recurrent-PREVALENT model 
structure is in Figure \ref{fig_recurrentprevalent}.

Before computing the prediction of actions, the model selects a set of candidate views from $O_i$. After that, the VLN$\circlearrowright$BERT %Recurrent PREVALENT 
model projects the candidate views and the instruction into the same feature space. We discuss this process in detail in Appendix A. Eventually, we have a vector of instruction features $F^{t=1}_{instruction} = [f_{cls},f_{w_1},..,f_{w_L},f_{sep}]$ and a vector of candidate action features $F^{t}_{candidate} =  [f_{a_1},...,f_{a_n},f_{a_{stop}}]$ as inputs of the action prediction.

At the first time step, $F^{t=1}_{instruction}$ is sent to a 9-layer self-attended module. The word features are thus attended to the $f^{t = 1}_{cls}$ feature. The model then appends $f^{t = 1}_{cls}$ to $F^{t=1}_{candidate}$ from $F^{t=1}_{instruction}$. After that, a cross-attention sub-module attends the remaining elements in $F^{t=1}_{instruction}$ to both $F^{t=1}_{candidate}$ and $f^{t = 1}_{cls}$. Lastly, another sub-module computes the self-attention of the instruction-attended $[F^{t=1}_{candidate},f^{t = 1}_{cls}]$. Such cross and self sub-modules build up the cross + self-attention module in figure \ref{fig_recurrentprevalent}. The process repeats for four layers and the attention scores between $f^{t=1}_{cls}$ and each elements in $F^{t=1}_{candidate}$ of the last layer are the prediction scores of each action $p_1,...,p_n,p_{stop}$. Additionally, the $f^{t}_{cls}$ in the output is sent to a cross-modal-matching module. The output of the module is used as $f^{t+1}_{cls}$ in the next time step while other features in $F^{t=1}_{instruction}$ remains unchanged. The cross and self attention computation will be repeated to compute action predictions for the rest of time steps. 

The VLN$\circlearrowright$BERT model minimizes two losses: imitation learning loss and reinforcement learning loss:
$$\mathcal{L}_{\textnormal{original}} = -\lambda\sum^{T}_{t=1}a_t\log(p_t)-\sum^{T}_{t=1}a_s\log(p_t)A(t)$$
where $a_t$ is the teacher action (one-hot encoded action that gets closest to the destination), $p_t$ is the probability of the taken action, $a_s$ is the action taken and $A(t)$ is the advantage value at time step $t$, computed by the A2C algorithm \cite{mnih2016asynchronous}. $\lambda = 0.5$ is a hyper-parameter that balances the weights of imitation learning loss and reinforcement learning loss.

\section{Proposed Method}
\subsection{Differences of Snapshots in the Same VLN model}
\label{differror}
Like other machine learning models, VLN$\circlearrowright$BERT chooses the best snapshot by validation to represent the trained model. %However, there's no guarantee the model always works as the best snapshot shows. Particularly, this is not true for the VLN model. 
We train the VLN$\circlearrowright$BERT model and observe its validation success rates, as measured on the val\_unseen split of R2R, of the snapshots saved in the training process. As shown in Appendix B Figure 1, we saw that the success rate fluctuates drastically over time. This fluctuation is however not seen in the training loss. As shown in Appendix B, figures 2 and 3, both imitation and reinforcement learning losses drop consistently with time. This interesting discovery leads us to further investigate whether the snapshots that perform similarly (in terms of success rates) might behave differently with respect to the errors that they make. %However, when observing the curve of imitation learning loss and reinforcement learning loss, the losses drop in a much more consistent way. Such observation leads to a further investigation: do these snapshots act differently while their performances are still close to each other?

%Thus, a further investigation is conducted on whether any of these snapshots %could ``well-represent" the overall performance of Recurrent PREVALENT. 
We set up an experiment designed as follows: we train the VLN$\circlearrowright$BERT model for 300,000 iterations and save the best snapshot in the validation split for every 30,000 iterations. The top-5 snapshots among them are shown in Table \ref{tbl_ensemble12vanilla}. We chose the best two snapshots, namely the snapshots with 62.32\% and 61.60\% success rates. We then count the navigations that only one of the snapshots failed, both of the snapshots failed or none of the snapshots failed. Our result shows that 563 navigations ended with different results between the best and the second-best snapshots, approximately 24\% of the validation data. In comparison, the difference in their success rate is only 0.72\%. The massive difference between 24\% and 0.72\% suggests that the agents of the two snapshots have different navigation behaviors even though they are almost equal in success rates. Naturally, we wonder if we could leverage both of their behaviors and thus create a better agent. One of the techniques we find to be effective for this problem is the snapshot ensemble that we discuss in the next section.

\begin{table}[ht]
\small
\begin{tabular}{|c|c|}
\hline
Snapshot Period & Success Rate in  val\_unseen Split \\ \hline

90K - 120K      & 62.32\%                             \\ \hline
240K - 270K     & 61.60\%                            \\ \hline
210K - 240K     & 61.56\%                              \\ \hline
60K - 90K       & 61.52\%                              \\ \hline
180K - 210K     & 61.30\%                              \\ \hline

\end{tabular}
\caption{The navigation success rates for the top-5 snapshots of VLN$\circlearrowright$BERT  in 10 periods of a 300,000-iteration training cycle.} 
\label{tbl_ensemble12vanilla}
\vspace{-1em}
\end{table}

% \begin{figure}[ht]
% \centering
% \includegraphics[width=1.0\columnwidth]{figures/SRCurve.png} 

% \caption{The curve of validation success rate over time during training. We can observe a drastic fluctuation throughout the training.} 
% \label{fig_SRcurve}
% \vspace{-1em}
% \end{figure}

\subsection{Snapshot Ensemble for VLN$\circlearrowright$BERT models}
A snapshot is a set of saved parameters of a model during a particular time in training. Naturally, the first thing to do to set up the ensemble is to decide what snapshots to save during training. According to \citet{huang2017snapshot}, the ensemble mechanic does the best when ``the individual models have low test error and do not overlap in the set of examples they misclassify''. Therefore, we want to save snapshots that are ``different enough" while doing well individually. Our approach is as follows (where snapshots and ensembles are evaluated on the validation set):

\begin{itemize}
\item For a training cycle of $N$ iterations, we evenly divide it into $M = \{m_1,m_2,...,m_{M}\}$ periods (assuming $N$ is divisible by $M$). 
\item For each period $m_i$, we save the snapshot $s_i$ with the highest success rate in the validation split.
\item The saved snapshots will be the candidates to build the ensemble: $\{s_1,...,s_{M}\}$.
\end{itemize}

Among the candidates snapshot saved this way, we conduct a beam search with size $l = 3$ to construct the ensemble of maximum size $k$. The process is as follows:

\begin{itemize}
\item Evaluate all possible ensembles of size 1, that is: $\{\{s_1\},\{s_2\},...,\{s_M\}\}$
\item Keep the top-$l$ ensembles in the previous step. For each kept ensemble, evaluate all possible ensembles of size 2 that contain the kept snapshot(s). E.g., say $\{s_2\}$ is one of the top-$l$ ensembles of size 1, the ensembles of size 2 to be evaluated related to $\{s_2\}$ are: $\{\{s_2,s_1\},\{s_2,s_3\},...,\{s_2,s_M\}\}$.
\item Keep the top-$l$ ensembles of size 2 from the previous step. Then we repeat the process for size-3 ensembles, so on and so forth. The evaluation stops when we finish evaluating the ensembles of size $k$.
\item In the end, we choose the ensemble with the highest success rate among all the ensembles evaluated during the whole process.
\end{itemize}

The approximate number of evaluations needed for our beam search strategy is $ O(Mlk)$ when $M >> k$, which is much smaller than the cost of an exhaustive search $O(min(M^k, M^{(M-k)})$.

\begin{figure}[ht]
\centering
\includegraphics[width=0.9\columnwidth]{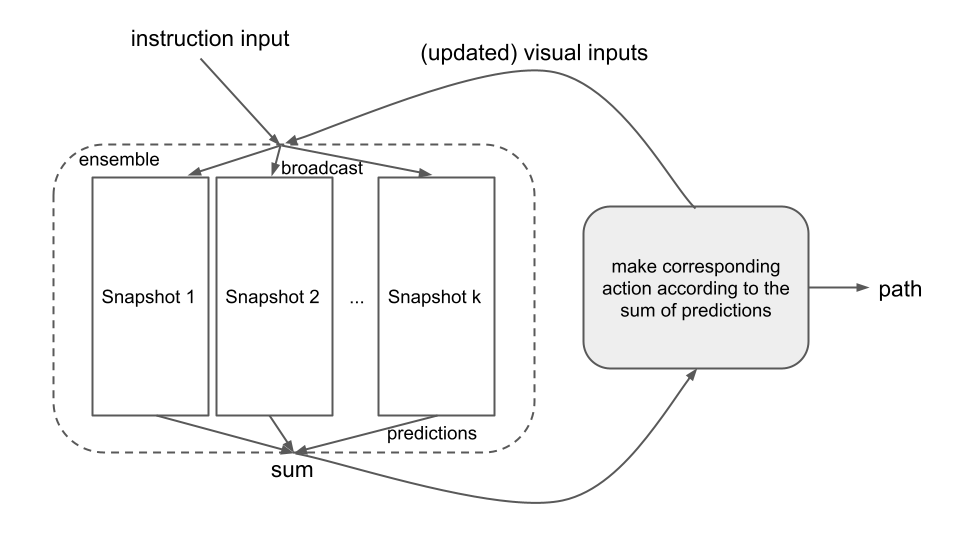} 
\caption{The workflow of the snapshot ensemble in the recurrent navigation process. The inputs broadcast to all snapshots, and the ensemble sums their predictions to make action at every time step.} 
\label{fig_ensemble_nav_workflow}
\vspace{-1em}
\end{figure}

During the evaluation, the ensemble completes a navigation task as follows: at each time step, the instruction inputs and the visual inputs of the current viewpoint are sent to each snapshot in the ensemble. Each snapshot then gives its predictions on the available actions. After that, the agent sums those predictions up and takes the corresponding action. At the end of the time step, each snapshot uses the action taken to update its own states. We visualize the ensemble navigation workflow in Figure \ref{fig_ensemble_nav_workflow}. We do not apply normalization on the prediction scores of snapshots to allow model confidence as the score weights. 

\subsection{A past-action-aware VLN$\circlearrowright$BERT model}

We saw a significant improvement in the snapshot ensemble of the original VLN$\circlearrowright$BERT model, as shown in Table \ref{evaluation_r2r}. Still, we could improve the performance of the ensemble by adding more variant snapshots. To do that, we modify the VLN$\circlearrowright$BERT model and combine the snapshots of the original and the modified model. 

\begin{figure}[ht]
\centering
\includegraphics[width=0.9\columnwidth]{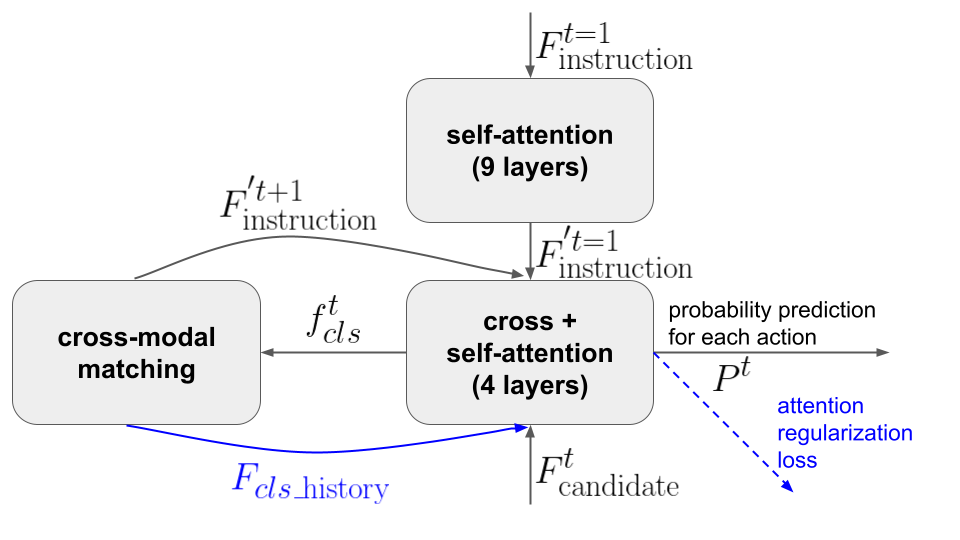}
\caption{The structure of past-action-aware VLN$\circlearrowright$BERT model. The blue parts are the modification added to the original structure. A $cls$ history vector is added to keep track of the past $f^{t}_{cls}$. An attention regularization loss is added to regularize the attention scores between $f^{t}_{cls}$ at each time step and relevant words in the instruction (visualization is given in appendix C).} 
\label{fig_pastactionaware_rprevalent}
\vspace{-1em}
\end{figure}

Our modification is based on the two ideas to improve the model: adding $f^{t}_{cls}$ in the past time steps and regularizing the attention scores between $f^{t}_{cls}$ and words in $I$ based on the observation of OSCAR-init VLN$\circlearrowright$BERT model in \cite{hong2021vln}. The modification is visualized by the blue parts in figure \ref{fig_pastactionaware_rprevalent}. 

At the beginning of each time step $t>1$, we add a copy of the cross-modal matching output $f^{'t}_{cls}$ from the last time step to a $F_{cls\_history}$ vector. At time step $t$, $F_{cls\_history} = [f^{'1}_{cls},...,f^{'t-1}_{cls}]$, re-indexed from $2$ to $t$. %To distinguish $f^{'t}_{cls}$ in $F_{cls\_history}$ from the $f^{'t}_{cls}$ in $F^{'t}_{candidate} + f^{'t}_{cls}$, we ordered the indexes in $F_{cls\_history}$ from $1$: $F_{cls\_history} = [f^{'1}_{cls},...,f^{'t-1}_{cls}]$. 
We then concatenate $F_{cls\_history}$ and [$F^{t}_{candidate},f^{'t}_{cls}$] as a large vector, and pass them through the cross + self-attention module. Note that we do not update the features in $F_{cls\_history}$ during the attention computation. As a result, we will not only have the attention scores from the current $f^{'t}_{cls}$ to each word features in $F^{'t=1}_{instruction}$, but also that from the $F_{cls\_history}$ to $F^{'t=1}_{instruction}$ in the last layer outputs.

For each set of attention scores $X_i = [x_1,...,x_L]$ from $f^{'i}_{cls}$ to each word in the instruction $w_1,..,w_L$ where $i \in [1,t]$, we compute an ``attention regularization" loss defined as follows:
$$\mathcal{L}_{\textnormal{attention}_t}= \frac{1}{t} \sum^{t}_{i=1}\textnormal{MSE}(\tanh (X_i), G_i)$$
``MSE" stands for Mean-Squared-Error and $G_i = [g_{i,1},...,g_{i,L}]$ is the ``ground truth" values for the normalized attention scores $\tanh (X_i)$. $G_i$ is computed based on the sub-instruction annotation from the Fine-Grained R2R dataset (FGR2R). Concretely, the FGR2R dataset divides the instructions in the R2R dataset into a set of ordered sub-instructions: $I = [I_{sub_{1}},I_{sub_{2}},...,I_{sub_{n}}]$ where n is the number of sub-instructions the original instruction consists of. Each sub-instruction corresponds to one or a sequence of viewpoints in the ground truth path $P = \{v_1,v_2,...,v_m\}$. To compute $G_i$, we first build a map from each viewpoint $v_i$ in $P$ to a specific sub-instruction in $I$. The map function is very straightforward: we choose the first sub-instruction $I_{sub_{i}}$ in $I$ that corresponds to $v_i$ as the mapped sub-instruction. By doing so, each viewpoint $v$ in $P$ now has their own related sub-instruction $I_{sub_{i}}$ in $I$. We then compute $G_i = [g_1,...,g_L]$, by the following step:
\begin{itemize}
\item find the viewpoint $v_i$ where the agent stands at time step $i$. If $v_i \notin P$, we choose the viewpoint in $P$ that is closest to $v_i$ as the new $v_i \in P$.
\item Since every $v_i$ has its mapped $I_{sub_{i}}$, we compute each $g_{j} \in G_i$ by: 
$$g_{j} = \left\{ 
  \begin{array}{ c l }
    1 & \quad \textrm{if } w_j  \in I_{sub_{i}}\textrm{,} \\
    0.5                 & \quad \textrm{if } w_j \in I_{sub_{i+1}}\textrm{,} \\
    -1                   & \quad \textrm{otherwise.}
  \end{array}
\right.$$

\end{itemize}
We compute each $\mathcal{L}_{\textnormal{attention}_t}$ and the total loss becomes: $$\mathcal{L}_{\textnormal{past-action-aware}} = \mathcal{L}_{\textnormal{original}} + \alpha\sum^{T}_{t=1}\mathcal{L}_{\textnormal{attention}_t}$$
$\alpha = 0.5$ is a hyper-parameter and $T$ is the total time steps.
% With the additional regularization loss, the total losses to minimize are: $$\mathcal{L}_{\textnormal{past-action-aware}} = \mathcal{L}_{\textnormal{original}} + \alpha\sum^{T}_{t=1}\mathcal{L}_{\textnormal{attention}_t}$$
% where $\alpha = 0.5$ is a hyper-parameter for weight balance.

The added $cls$ history vector provides additional information to the model during action prediction. In addition, the attention regularization forces the VLN$\circlearrowright$BERT model to align attention scores to words that correspond to the agent's actions without performance (success rate) drop. The visualization of how the attention score changes as the agent moves in its path are given in appendix C. 

\begin{table*}[ht]
\small
\centering
\small
\begin{tabular}{|c|cccc|cccc|} 
\hline
\multirow{2}{*}{Model}                               & \multicolumn{4}{c|}{R2R val\_unseen} & \multicolumn{4}{c|}{R2R test}     \\ 
\cline{2-9}
                                                     & TL$\downarrow$    & NE$\downarrow$   & SR$\uparrow$    & SPL$\uparrow$        & TL$\downarrow$    & NE$\downarrow$   & SR$\uparrow$     & SPL$\uparrow$     \\ 
\hhline{|=========|}
Random                                               & 9.77  & 9.23 & 16    & -         & 9.89  & 9.79 & 13    & 12     \\
Human                                                & -     & -    & -     & -         & 11.85 & 1.61 & 86    & 76     \\ 
\hline
Seq2Seq-SF \cite{anderson2018vision}                                           & 8.39  & 7.81 & 22    & -         & 8.13  & 7.85 & 20    & 18     \\
Speaker-Follower \cite{fried2018speaker}                                   & -     & 6.62 & 35    & -         & 14.82 & 6.62 & 35    & 28     \\
PRESS \cite{li2019robust}                                                & 10.36 & 5.28 & 49    & 45        & 10.77 & 5.49 & 49    & 45     \\
EnvDrop \cite{tan2019learning}                                             & 10.7  & 5.22 & 52    & 48        & 11.66 & 5.23 & 51    & 47     \\
AuxRN   \cite{Zhu_2020_CVPR}                                             & -     & 5.28 & 55    & 50        & -     & 5.15 & 55    & 51     \\
PREVALENT \cite{hao2020towards}                                           & 10.19 & 4.71 & 58    & 53        & 10.51 & 5.3  & 54    & 51     \\
RelGraph  \cite{hong2020language}                                           & 9.99  & 4.73 & 57    & 53        & 10.29 & 4.75 & 55    & 52     \\ 
\hline
VLN$\circlearrowright$BERT \cite{hong2021vln}                                 & 12.01 & 3.93 & 63    & 57        & 12.35 & 4.09 & 63    & 57     \\
VLN$\circlearrowright$BERT + REM \cite{liu2021vision}                          & 12.44 & 3.89 & 63.6  & 57.9      & 13.11 & 3.87 & \textbf{65.2}  & 59.1   \\
\hline
Past-action-aware VLN$\circlearrowright$BERT (ours)         & 13.2  & 3.88 & 63.47 & 56.27     & 13.86 & 4.11 & 62.49 & 56.11  \\
VLN$\circlearrowright$BERT Original Snapshot Ensemble (ours) & 11.79 & 3.75 & 65.55 & 59.2      & 12.41 & 4    & 64.22 & 58.96  \\
VLN$\circlearrowright$BERT Past-action-aware Snapshot Ensemble (ours) & 12.35 & 3.72 & 65.26 & 58.65      & 13.19 & 3.93    & 64.65 & 58.78  \\
VLN$\circlearrowright$BERT Mixed Snapshot Ensemble (ours)   & 12.05 & \textbf{3.63} & \textbf{66.67}  & \textbf{60.16}     & 12.71 & \textbf{3.82} & 65.11 & \textbf{59.61}  \\
\hline
\end{tabular}
\caption{The evaluation results for our snapshot ensemble and past-action-aware VLN$\circlearrowright$BERT models (bold is best). Our mixed snapshot ensemble achieved the new SOTA performance in NE and SPL, and only 0.09\% worse than \cite{liu2021vision}, which uses further data augmentation, in SR.}
\label{evaluation_r2r}
\end{table*}

\begin{table*}
\small
\centering
\begin{tabular}{|c|cccc|cccc|} 
\hline
\multirow{2}{*}{Model}                               & \multicolumn{4}{c|}{R4R val\_unseen\_half} & \multicolumn{4}{c|}{R4R val\_unseen\_full}  \\ 
\cline{2-9}
                                                     & TL$\downarrow$    & NE$\downarrow$   & SR$\uparrow$    & SPL$\uparrow$        & TL$\downarrow$    & NE$\downarrow$   & SR$\uparrow$     & SPL$\uparrow$                   \\ 
\hhline{|=========|}
Speaker-Follower                                     & -  & -  & -     & -                        & 19.9 & 8.47 & 23.8  & 12.2                  \\
EnvDrop                                              & -  & -  & -     & -                        & -    & 9.18 & 34.7  & 21                    \\
VLN$\circlearrowright$BERT + REM  \cite{liu2021vision}                        & -  & -  & -     & -                        & -    & \textbf{6.21} & \textbf{46}    & \textbf{38.1}                  \\ 
\hline
VLN$\circlearrowright$BERT                                & 13.76  & 7.05  & 37.29 & 27.38                        & 13.92    & 6.55    & 43.11 & 32.13                     \\
VLN$\circlearrowright$BERT Original Snapshot Ensemble (ours) & 15.09  & \textbf{7.03}  & \textbf{39}     & \textbf{28.66}                        & 14.71    & 6.44  & 44.55 & 33.45                     \\
\hline
\end{tabular}
\caption {Evaluation results on R4R dataset (bold is best). We also present the evaluation result of the full split with our constructed ensemble. The model gains $>1\%$ improvement from the original VLN$\circlearrowright$BERT model after applying snapshot ensemble. Note that \cite{liu2021vision} uses further data augmentation, which is orthogonal to our approach. At the time of writing they have yet to release their code or dataset which might improve our performance further.}
\label{R4R_val_unseen}
\vspace{-1em}
\end{table*}

\begin{table*}[htp]
\small
\centering
\arrayrulecolor{black}
\begin{tabular}{|c|cccc|cccc|} 
\hline
\multirow{2}{*}{Model}            & \multicolumn{4}{c|}{R2R val\_unseen} & \multicolumn{4}{c|}{R2R Test}  \\ 
\cline{2-9}
                                  & TL$\downarrow$ & NE$\downarrow$ & SR$\uparrow$     & SPL$\uparrow$           & TL$\downarrow$ & NE$\downarrow$ & SR$\uparrow$     & SPL$\uparrow$     \\ 
\hhline{|=========|}
EnvDrop                           & \textbf{10.7}  & 5.22 & 52    & 48            & \textbf{11.66} & 5.23 & 51    & 47      \\ 
\arrayrulecolor{black}
EnvDrop Snapshot Ensemble         & 11.74 & \textbf{4.9}  & \textbf{53.34} & \textbf{49.49}         & 11.9  & \textbf{4.98} & \textbf{53.58} & \textbf{50.01}   \\ 
\arrayrulecolor{black}\hline
OSCAR-init VLN$\circlearrowright$BERT                  & \textbf{11.86} & 4.29 & 59    & 53            & 12.34 & 4.59 & 57    & 53      \\ 
\arrayrulecolor{black}
OSCAR-init VLN$\circlearrowright$BERT Snapshot Ensemble & 11.22 & \textbf{4.21} & \textbf{59.73} & \textbf{54.76}         & \textbf{11.74} & \textbf{4.36} & \textbf{59.72} & \textbf{55.35}   \\
\arrayrulecolor{black}\hline
\end{tabular}
\caption {The evaluation result of snapshot ensembles of EnvDrop and OSCAR-init VLN$\circlearrowright$BERT for R2R val\_unseen split. Both models are consistently improved by the snapshot ensemble methods.}
\label{evaluation_different_models}
\vspace{-1em}
\end{table*}

\section{Experiment}
We run the following experiments to evaluate the performances of snapshot ensembles in different models and datasets:
\begin{itemize}
\item We evaluate the performance of snapshot ensemble on the R2R dataset, including the ensemble built from the original model snapshots and the ensemble built from both the original and the modified model snapshots.
\item We also apply snapshot ensemble on the OSCAR-init VLN$\circlearrowright$BERT model from \citet{hong2021vln} and the Env-Drop model from \citet{tan2019learning}. We evaluate and compare their ensemble performances on the R2R dataset against their best single snapshot.
\item We evaluate the performance of snapshot ensemble on the R4R dataset, which is a larger VLN dataset than R2R and with more complicated navigation paths.
\end{itemize}
\subsection{Dataset Setting and Evaluation Metrics}
We use the R2R train split as training data, val\_unseen split as validation data, and test split to evaluate the ensemble. For the R4R dataset, we also use the train split as the training data. As there's no test split in the R4R dataset, we divide the val\_unseen split into two halves. The two halves do not share scenes in common. We construct the snapshot ensemble on one half and evaluate it on the other half. 

We adopt four metrics for evaluation: Success Rate (SR), Trajectory Length (TL), NavigationError (NE), and Success weighted by Path Length (SPL). SR is the ratio of successful navigation numbers to the number of all navigations (higher is better).
TL is the average length of the model's navigation path (lower is better). 
NE is the average distance between the last viewpoint in the predicted path and the ground truth destination viewpoint (lower is better); SPL is the path-length weighted success rate compared to SR (higher is better).
\subsection{Training Setting and Hard/Software Setup}
We train the VLN$\circlearrowright$BERT and the OSCAR-init VLN$\circlearrowright$BERT model with the default 300,000 iterations. We run an ablation study to decide $M = 10, k = 4$ for constructing the ensemble (the candidate number for beam search is $2M$ when mixing the original and modified models. In R4R, we set $k = 3$ to shorten the evaluation time).  Appendix D describes the ablation study result. For other parameters, we use the default given by the authors.\footnote{We do not adopt the cyclic learning rate schedule \cite{loshchilov2016sgdr} suggested in \citet{huang2017snapshot} that forces the model to generate local minima. Our trial experiment result shows there's no significant improvement by doing so in this task.}

We set the pseudo-random seed to 0 for the training process. We run the training code under Ubuntu 20.04.1 LTS operating system, GeForce RTX 3090 Graphics Card with 24GB memory. It takes around 10,000 MB of graphics card memory to evaluate an ensemble of 4 snapshots with batch size 8 inputs. The code is developed in Pytorch 1.7.1, and CUDA 11.2. The training takes approximately 30 - 40 hours to finish. The beam search evaluation is done in 3 - 5 hours.

\section{Results}
\subsection{Evaluation Results of our proposed methods on R2R dataset}
We evaluate our modification of the VLN$\circlearrowright$BERT model, the snapshot ensembles of original-only and mixed (original and modified) snapshots on the R2R test split. Table \ref{evaluation_r2r} shows the evaluation results of our model and ensembles. The past-action-aware modified model has a similar performance to the original model. Both snapshot ensembles significantly improve the performance of the model in NE, SR, and SPL. Specifically, the mixed snapshot ensemble achieved the new SOTA in NE and SPL while only 0.09\% worse than VLN$\circlearrowright$BERT + REM \cite{liu2021vision}, which uses further data augmentation, in SR. At the time of writing, they have yet to release their code or dataset which may improve our results further.

Additionally, we evaluate whether the snapshot ensemble also improves different VLN models. We train an OSCAR-init VLN$\circlearrowright$BERT \cite{hong2021vln}, which is a BERT-structure model, and an EnvDropout model \cite{tan2019learning} which has an LSTM plus soft-attention structure with their default training settings. We apply the snapshot ensemble on both and compare the ensemble's performance with their best single snapshot on the R2R test split. Table \ref{evaluation_different_models} shows the evaluation result. Both ensembles consistently gained a more than 2\% increase in SR and SPL compared to the best snapshot of the model. That suggests the snapshot ensemble is also able to improve the performances of other VLN models as well.

\subsection{Evaluation Results of Snapshot Ensemble on R4R dataset}

In addition to the R2R dataset, we evaluate the snapshot ensemble method on a more challenging dataset R4R \cite{jain2019stay}. R4R dataset contains more navigation data and more complicated paths in more variant lengths. Table \ref{R4R_val_unseen} shows the evaluation result between the best snapshot and the snapshot ensemble. We saw a more than 1\% of increase in SR and SPL after applying snapshot ensemble. We also evaluate the same snapshot ensemble with all val\_unseen split data in R4R, as shown in Table \ref{R4R_val_unseen}.

\section {Discussion}

In this section, we discuss potential reasons why the snapshot ensemble works well in the VLN task. Additionally, we provide a case study in appendix E that qualitatively analyzes our proposed snapshot ensemble.

\subsection{Ensemble is More Similar to Its Snapshots}
\label{simensemble}
\begin{figure}[ht]
\centering
\includegraphics[width=0.8\columnwidth]{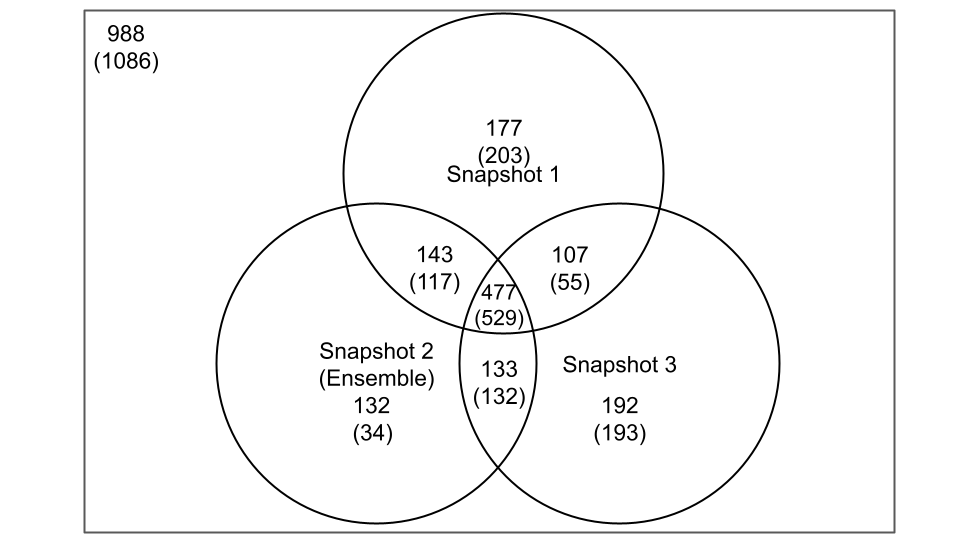}
\caption{The Venn diagram on val\_unseen that counts the number of navigation that are failed by one or more snapshots. The numbers not in any circle are navigations that are succeeded by all 3 snapshots. The numbers in parenthesis are the counts when snapshot 2 is replaced by the ensemble.} 
\label{fig_vennens3}
\vspace{-1em}
\end{figure}
To find out if the snapshot ensemble leverages the predictions of its snapshots, we analyze the errors that it makes in comparison to those of its snapshots (similar to our analysis in section \ref{differror}), counting the distinctive failures of the snapshots and the ensemble on the validation set. 
%We want to know if the snapshot ensemble takes advantage of its snapshots. Therefore, we count the distinctive failures of the snapshots for each navigation of the validation set like what we did in section 3.3. 
We choose an ensemble of 3 snapshots with SR 65.43\% obtained via beam search to visualize the shared failure cases among the three snapshots and the failure cases shared between snapshots 1, 3, and the ensemble (replacing snapshot 2). %, the best performing snapshot among the three (62.32\%), with the ensemble in the analysis). %we replace snapshot 2, which is the best snapshot among the three (62.32\%), with the ensemble and count the failure cases among the three models again.
We draw the corresponding Venn diagram with and without the replacement, as shown in Figure \ref{fig_vennens3}. 
In comparison to snapshot 2, the ensemble shares more navigations that are succeeded/failed between itself and the two snapshots (529+1086 $>$ 477+ 988). Meanwhile, the number of navigations that are only failed by the ensemble is less than that of snapshot 2 ($34 < 132$). These changes suggest that the ensemble behaves more similarly to its snapshot members than the replaced snapshot. We repeat this process by replacing snapshots 1 and 3 and obtain a similar result. 

To understand the benefits of the ensemble acting more similarly to its snapshots, we count the successful navigations of the ensemble/snapshots in each scene. Table \ref{tbl_scene_advs} shows the result. We find that each snapshot is good at different scenes. The ensemble either outperforms its snapshots or is comparable to %to be only one navigation behind 
the best snapshot in most scenes suggesting that the ensemble leverages the advantages of snapshots in different scenes to achieve better performance.

\subsection{Ensemble Avoids Long Navigations}
In our setting, the system forces the agent to stop when its taken actions are over 15. We call navigations that take the agent 15 or more actions to complete as Long Navigations (LNs). As ground truth navigation only needs 5 - 7 actions, we wonder if LN is harmful to the model performance. In table \ref{tbl_long_nav}, we count the LNs for snapshots and the ensemble discussed in section \ref{simensemble} and compute the success rates when their navigation is an LN. We discovered that 5.5\% to 10.5\% of agent's navigations are LNs. Meanwhile, LN has a high likelihood ($>$ 90\%) of failing. As the ensemble has a much fewer number of LNs than its snapshots (131), we consider avoiding more LNs as one of the reasons why the ensemble outperforms single snapshots.

\begin{table}[h]
\small
\centering
\arrayrulecolor{black}
\begin{tabular}{!{\color{black}\vrule}l!{\color{black}\vrule}l!{\color{black}\vrule}l!{\color{black}\vrule}l!{\color{black}\vrule}l!{\color{black}\vrule}} 
\arrayrulecolor{black}\hline
Scene    & Ensemble          & Snapshot 1  & Snapshot 2   & Snapshot 3    \\ 
\hline
1 & 178 & 165    & 169          & 159     \\ 
\arrayrulecolor{black}\hline
2 & 32           & 33 & 32      & 29       \\ 
\hline
3 & 140         & 131   & 131    & 144  \\ 
\hline
4 & 208 & 189   & 199          & 185     \\ 
\hline
5 & 10           & 11 & 8       & 9        \\ 
\hline
6 & 169          & 161    & 170 & 152     \\ 
\hline
7 & 203 & 198    & 200          & 196      \\ 
\hline
8 & 217 & 205    & 204     & 212           \\ 
\hline
9 & 93  & 80     & 89           & 84       \\ 
\hline
10 & 102 & 95          & 89      & 89       \\ 
\hline
11 & 185 & 177    & 173     & 181           \\
\arrayrulecolor{black}\hline
\end{tabular}
\caption {The count of successful navigations for the ensemble and its snapshots in each scene on val\_unseen split.}
\label{tbl_scene_advs}
\vspace{-1em}
\end{table}

\begin{table}[ht]
\small
\centering
\arrayrulecolor{black}
\begin{tabular}{|l|r|r|r|} 
\hline
           & \multicolumn{1}{l|}{SR} & \multicolumn{1}{l|}{LN Count} & \multicolumn{1}{l|}{LN that fails (\%)
}  \\ 
\hline
Snapshot 1 & 61.52                   & 172                           & 159 (92.44\%)                                 \\ 
\arrayrulecolor{black}\hline
Snapshot 2 & 62.32                   & 155                           & 141 (90.97\%)                                 \\ 
\hline
Snapshot 3 & 61.3                    & 246                           & 223 (90.65\%)                                 \\ 
\hline
Ensemble   & 65.43                   & 131                           & 123 (93.89\%)                                 \\
\arrayrulecolor{black}\hline
\end{tabular}
\caption {Long navigation (LN) for the ensemble and snapshots. The ensemble has fewer long navigation (131) when compared to its snapshot members.}
\label{tbl_long_nav}
\vspace{-1em}
\end{table}

\section {Conclusion}
In this paper, we discover differences in snapshots of the same VLN model. We apply the snapshot ensemble method that leverages the behaviors of multiple snapshots. By combining snapshots of the VLN$\circlearrowright$BERT model and its past-action-aware modification that we propose, we achieve a new SOTA performance on the R2R dataset. We also show that our snapshot ensemble method works with different models and on more complicated VLN tasks. In the future, we will train the model with augmented data from \citet{liu2021vision} and see if it improves performance. We will also apply our approach to pre-explore and beam search settings to see if ensemble methods can improve performance on these settings.

\bibliography{References.bib}
\end{document}

\maketitle

\section{Appendix A: Pre-processing of Recurrent PREVALENT model}

Recurrent PREVALENT does not process panoramic view $O_i$ directly. Instead, it uses the image feature of the view $O_{i,j}$ that leads to a particular navigable viewpoint to represent the action it goes there next. In addition, an all-zero image feature will be added to represent the stop action. In the other words, the visual inputs are the features of candidate actions: $O_{candidate} = [O_{a_1},...,O_{a_n},O_{a_{stop}}], O_{candidate} \in \mathbb{R}^{(n+1)\times k_1}$, where $n$ is the number of navigable viewpoints to go next and $k_1$ is the size of image feature.

On the other hand, Recurrent PREVALENT preprocesses the instruction inputs with pre-trained BERT embedding. Thus, $I = [w_{cls},w_1,...,w_L,w_{sep}], I \in \mathbb{R}^{(L+2)\times k_2}$ where $L$ is the instruction length and $k_2$ is the word embedding feature size. The $cls$ token in $I$ will be used as a hidden state for the recurrent process later.

To allow cross-attention between texts and images, Recurrent PREVALENT then projects the word and image features into a common feature space of size $k_3$: $$F_{instruction} = IW_{instruction} = [f_{cls},f_{w_1},..,f_{w_L},f_{sep}]$$ 
$$F_{candidate} = O_{input}W_{candidate} = [f_{a_1},...,f_{a_n},f_{a_{stop}}] $$ $W_{instruction} \in \mathbb{R}^{k1 \times k3}$ and $W_{candidate} \in \mathbb{R}^{k2 \times k3}$ are learnable weight matrices.

On the other hand, Recurrent PREVALENT preprocesses the instruction inputs with pre-trained BERT embedding. Thus, $I = [w_{cls},w_1,...,w_L,w_{sep}], I \in \mathbb{R}^{(L+2)\times k_2}$ where $L$ is the instruction length and $k_2$ is the word embedding feature size. The $cls$ token in $I$ will be used as a hidden state for the recurrent process later.

To allow cross-attention between texts and images, Recurrent PREVALENT then projects the word and image features into a common feature space of size $k_3$: $$F_{instruction} = IW_{instruction} = [f_{cls},f_{w_1},..,f_{w_L},f_{sep}]$$ 
$$F_{candidate} = O_{input}W_{candidate} = [f_{a_1},...,f_{a_n},f_{a_{stop}}] $$ $W_{instruction} \in \mathbb{R}^{k1 \times k3}$ and $W_{candidate} \in \mathbb{R}^{k2 \times k3}$ are learnable weight matrices.

\section {Appendix B: Curves of Validation Success Rate, Imitation Learning and Reinforcement Learning Losses Over Time during Training}
\begin{figure}[h]
\centering
\includegraphics[width=1.0\columnwidth]{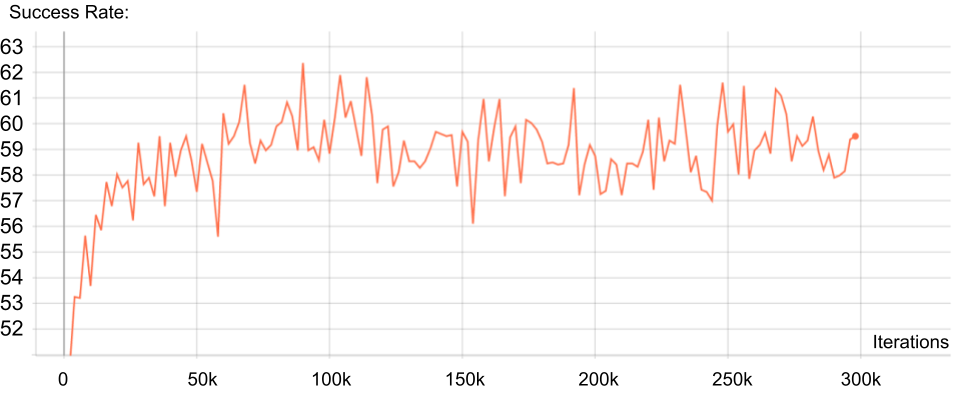} 
\caption{The curve of validation success rate over time during training. We can observe a drastic fluctuation throughout the training.} 
\label{fig_SRcurve}
\end{figure}

\begin{figure}[h]
\centering
\includegraphics[width=1.0\columnwidth]{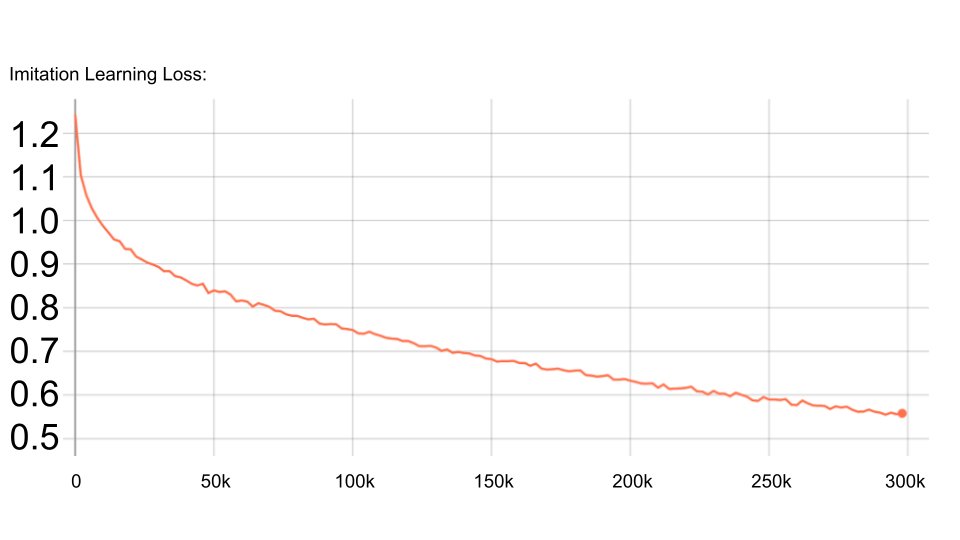} 
\caption{The curve of imitation learning loss. We can observe a consistent drop throughout the training.} 
\label{fig_ILcurve}
\end{figure}

\begin{figure}[h]
\centering
\includegraphics[width=1.0\columnwidth]{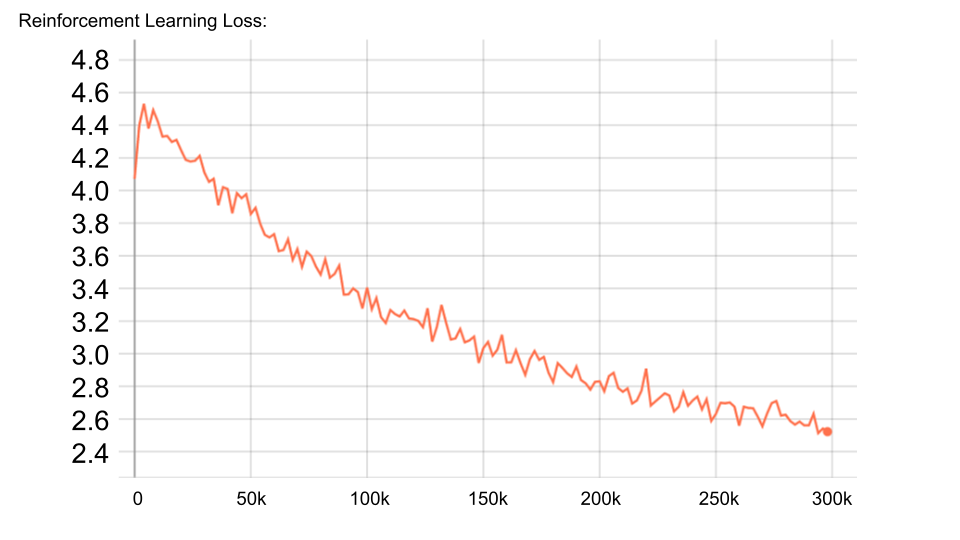} 
\caption{The curve of reinforcement learning loss. We can also see a consistent drop throughout the training.} 
\label{fig_RLcurve}
\end{figure}

Figure \ref{fig_SRcurve} shows the curve of validation success rate over time during training the original Recurrent PREVALENT model. Figure \ref{fig_ILcurve} and figure \ref{fig_RLcurve} show the imitation learning and reinforcement learning loss curves throughout the training process. The training setting is the same as described in section 5.2.

\section {Appendix C: Visualization of the Attention Scores Before/After Regularization}

\begin{figure*}[ht]
\centering
\includegraphics[width=1.5\columnwidth]{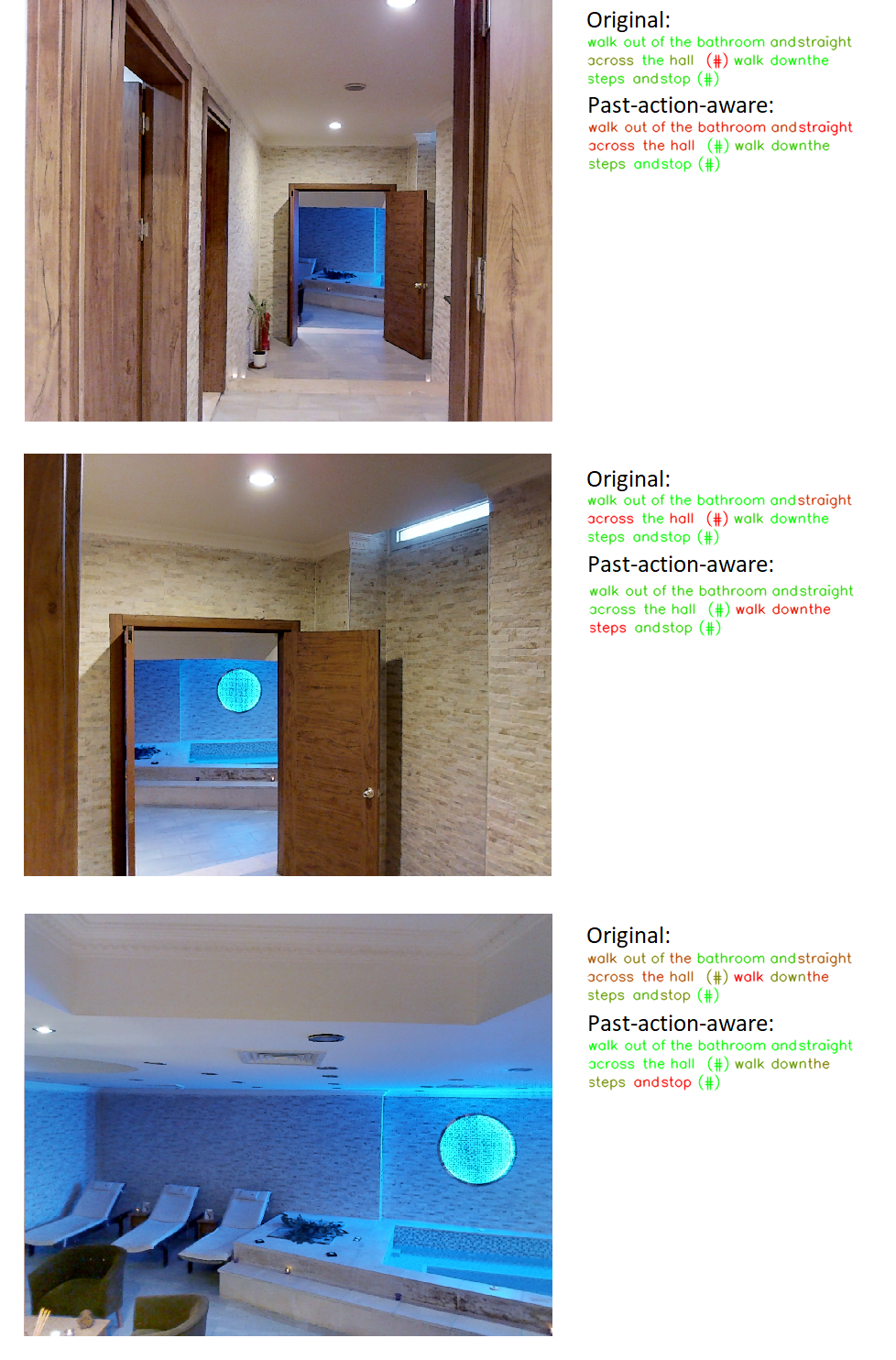} % Reduce the figure size so that it is slightly narrower than the column. Don't use precise values for figure width.This setup will avoid overfull boxes.
\caption{The visualization of attention scores between the $f^{t}_{cls}$ and each word in the $F^{t}_{instruction}$ vector from three different time steps. The greener the word is, the lower attention score it has. The more red the word is, the higher the attention score the word has. For better visualization, we replace period ``." with ``(\#)" in the figure. We can see a more concentrated red word area in the attention scores from the past-action-aware modification model. The red word areas from the action-aware-attention one also make more logical sense as ``sub-instructions" to be focused on.} 
\label{fig_past_action_aware}
\end{figure*}

We visualize the regularized attentions scores between $f^{t}_{cls}$ and each word in the instruction feature vector $F^{t}_{instruction}$. And we also visualize the unregularized one as a comparison. The visualization is shown in figure \ref{fig_past_action_aware}. We observe that the past-action-aware modification has a more concentrated high attention word area. The red word area in the past-action-aware modification also makes more logical sense as the part of the instruction the agent focuses on.

\section {Appendix D: the Ablation Study for $M$ and $k$ of Snapshot Ensemble}
To find out the influences of $M$ and $k$ on the performance of snapshot ensemble, we evaluate the performances of snapshot ensembles with different $M$ and $k$'s in the R2R val\_unseen split. We fixed $k=3$ and $M = 10$ as the initial setting for the ablation study experiments. We tested $M$ from $\{5,10,15\}$ and $k$ from $\{3,4,5\}$. The results are shown in table \ref{tbl_ablation_m} and table \ref{tbl_ablation_k}.

\begin{table}[h]
\centering
\begin{tabular}{|c|cccc|} 
\hline
M  & TL$\downarrow$    & NE$\downarrow$   & SR$\uparrow$    & SPL$\uparrow$   \\ 
\hline
5  & 12    & 3.87 & 64.58 & 58.32  \\
10 & 11.79 & 3.77 & 65.18 & 58.88  \\
15 & 12.08 & 3.73 & 65.3  & 58.88  \\
\hline
\end{tabular}
\caption{The ablation study experiment for the number of snapshots to save $M$. Here we fixed $k = 3$. We saw significant improvement when $M$ increases from 5 to 10 (0.40\% in SR) but a minor improvement from 10 to 15 (0.12\% in SR).}
\label{tbl_ablation_m}
\end{table}

\begin{table}[h]
\centering
\begin{tabular}{|c|cccc|} 
\hline
M  & TL$\downarrow$    & NE$\downarrow$   & SR$\uparrow$    & SPL$\uparrow$    \\ 
\hline
3 & 11.79 & 3.77 & 65.18 & 58.88  \\
4 & 11.8  & 3.75 & 65.56 & 59.2   \\
5 & 11.88 & 3.8  & 65.69 & 59.36  \\
\hline
\end{tabular}
\caption{The ablation study experiment for the maximum number of snapshots to be in the ensemble $k$. Here we fixed $M=10$. We saw an increase of SR when $k$ increases from 3 to 4  (0.38\%) but not that much from 4 to 5 (0.13\%).}
\label{tbl_ablation_k}
\end{table}

According to the results, we find a significant increase of SR from $M = 5$ to $M = 10$ (0.40\%), while not that much (0.12\%) from $M = 10$ to $M = 15$. Considering $M = 15$ takes 50\% more ensembles to evaluate, we choose $M = 10$ to be our number of snapshots to save during training.

After fixing $M = 10$, we discover that the ensemble performance improves by 0.38\% when $k$ increases from 3 to 4. A much less improvement is seen when $k$ increases from 4 to 5 (0.13\%). Since setting $k = 5$ requires another 3,000 MB graphics card memory and extra sets of ensembles for evaluation but with no significant improvement, we decide to choose $k = 4$ as our number of maximum snapshots in the ensemble during beam search.

\section {Appendix E: Case Study for Recurrent PREVALENT Snapshot Ensemble}
\begin{figure*}[ht]
\centering
\includegraphics[width=1.9\columnwidth]{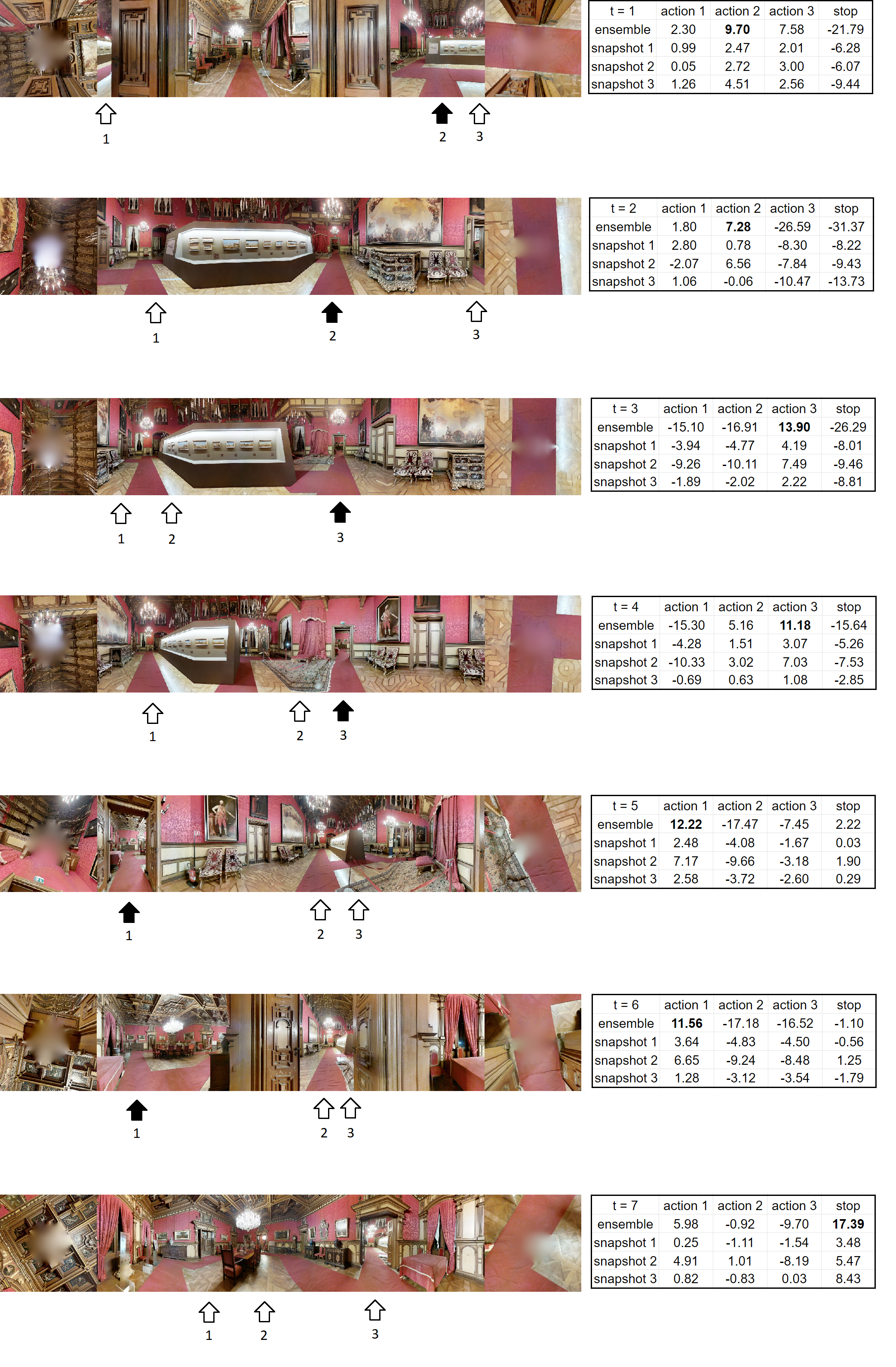} % Reduce the figure size so that it is slightly narrower than the column. Don't use precise values for figure width.This setup will avoid overfull boxes.
\caption{The visualization of the prediction scores of the ensemble and its member snapshots during navigation. The table on the right shows the prediction scores of each snapshot taking action 1, 2, 3, or stop in the current time step. The arrows below the panoramic view image point out the directions of each action goes.} 
\label{fig_case_study}
\end{figure*}

Figure \ref{fig_case_study} visualizes a case of our snapshot ensemble agent finishing navigation in a museum-like environment. The instruction is ``Go through the large wooden doors and turn right. Pass the photos on the left and pass the second set of wooden doors. Continue going straight and stop at the chair at the end of the table." In most time steps, we can see that all snapshots contribute to deciding what the ensemble should act next. However, exceptions exist. In time step $t = 2$, snapshots 1 and 3 both ignored ``turn right" and voted to take action 1. As the only correct snapshot among three, snapshot 2 ``forced" the ensemble to take action 2 by predicting the action with a much higher prediction score. The above observation suggests such a weighted voting mechanic helps improve the ensemble performance from its member snapshots.

% \section {Acknowledgement}
% Acknowledgement. %%%%Terry-san: please let me know what needs to be written here according to Honda's rules. %%%%
